\pdfoutput=1
\documentclass[11pt]{article}

\usepackage[preprint]{acl}
\usepackage{times}
\usepackage{latexsym}
\usepackage[T1]{fontenc}
\usepackage[utf8]{inputenc}
\usepackage{microtype}
\usepackage{inconsolata}
\usepackage{graphicx}
\usepackage{booktabs}
\usepackage{amsmath}
\usepackage{mathtools}
\usepackage{multirow}
\usepackage{subcaption}
\usepackage{xcolor}
\usepackage{url}
\usepackage{hyperref}
\usepackage{mdframed}

\hypersetup{
  pdftitle={When Does Span-Guided Detoxification Help? Human Preferences and Evaluator Diagnostics in a Controlled Comparison},
  pdfauthor={Kyungwon Park}
}

\title{When Does Span-Guided Detoxification Help?\\
Human Preferences and Evaluator Diagnostics in a Controlled Comparison}

\author{
Author Name(s)\\
Affiliation(s)\\
\texttt{email@example.com}
}
\author{
  Kyungwon Park\\
  Department of Artificial Intelligence, Yonsei University\\
  \texttt{\ cosmicboon@yonsei.ac.kr}
}

\begin{document}
\maketitle

\begin{abstract}
Span-guided rewriting aims to preserve meaning by localizing edits to annotated harmful spans, but the same constraint can leave harmful intent insufficiently mitigated. We present a controlled exploratory comparison of span-guided and unguided detoxification on a mixed-source English evaluation set comprising manually curated inputs and HateXplain test items. We conduct a dense blinded human evaluation under a fixed single-generator setting.

Human preferences reveal a trade-off rather than a uniformly superior rewriting strategy. Span-guided outputs are favored when localized editing preserves the original stance and avoids unnecessary modification, whereas unguided outputs are favored when broader rewriting achieves more complete mitigation. This contrast varies substantially across the study-defined strata: the two strategies are competitive in the strong stratum, while unguided rewriting is clearly preferred in the mild stratum. Rationale annotations trace this difference to complementary failure risks---residual harm after localized editing and over-modification after broader rewriting.

We treat automatic evaluation as a diagnostic rather than a substitute for human judgment. Toxicity--similarity scalarizations, a multi-generator analysis, and two general-purpose LLM judges reproduce parts of the aggregate tendency but do not yield an analogous stratified contrast. These setting-specific findings do not establish a severity-based routing rule. Instead, they motivate evaluation protocols that assess mitigation sufficiency and meaning preservation separately and report both residual harm and over-modification alongside aggregate scores.
\end{abstract}

\section{Introduction}

Large language models are increasingly used in interactive systems, content production, and moderation workflows, creating demand for methods that can rewrite harmful language without unnecessarily changing the speaker's intended content~\citep{gehman2020realtoxicityprompts}. Potential applications include human-in-the-loop moderation assistance, authoring-time toxicity reduction, and post-generation safety editing. In each setting, a useful rewrite must balance two requirements: harmful content should be mitigated sufficiently, and non-harmful meaning or pragmatic intent should be preserved.

Text detoxification has therefore been studied through parallel rewriting, prompting, controlled generation, and decoding-time interventions~\citep{logacheva2022paradetox,agarwal2023haterephrase,gedi,dexperts,khondaker2024detoxllm,lee2024xdetox}. A common design choice is whether to localize edits around detected toxic spans or allow a model to reformulate the sentence more broadly. Span-guided rewriting can reduce unnecessary changes, but a narrow edit may retain the harmful proposition or pragmatic attack. Unguided rewriting has more freedom to mitigate the message, but may over-sanitize, distort stance, or introduce a new discourse frame.

Evaluation usually combines toxicity reduction with semantic preservation, sometimes through a single weighted score~\citep{toshevska2021review,jin2022deep,khondaker2024detoxllm,lee2024xdetox}. These metrics are useful diagnostics, but they do not directly specify how much mitigation is sufficient for a particular input or when preservation becomes excessive. Prior work has documented disagreement between automatic metrics and human judgments in text generation evaluation~\citep{deriu2023correction}; the remaining question is how the competing risks of under-mitigation and over-modification appear in a controlled strategy comparison.

We study two concrete prompting conditions: an unguided condition that permits whole-sentence rewriting and a span-guided condition that supplies item-level harmful spans and requests localized editing. For the HateXplain-derived items, these spans come from the dataset's gold rationale annotations. The generator and decoding settings are fixed in the human study. Importantly, this is a \emph{bundled} prompt contrast: the conditions differ both in span information and in the instruction about edit locality. We therefore characterize the behavior of these two implementations rather than claiming an isolated causal effect of span availability.

Our primary evidence is a dense human evaluation of 60 English inputs balanced across two study-defined operational strata (strong and mild). The set combines 30 manually curated items with 30 HateXplain test items; these strata are analytical groupings used in the evaluation corpus, not a validated or universal harm-severity taxonomy. Automatic toxicity and similarity measures, a larger five-generator study, and two LLM judges are reported as secondary evaluator diagnostics.

We make three contributions:
\begin{enumerate}
  \item We provide a controlled human preference study showing that localized and global rewriting expose distinct failure risks: residual harm after narrow edits and over-modification after broader rewrites.
  \item We show a large association between the study-defined evaluation stratum and item-level strategy preference, while explicitly avoiding a causal or universal severity interpretation.
  \item We separate human evidence from evaluator diagnostics: scalar-proxy analyses favor unguided rewriting in their evaluated HateXplain samples, while two general-purpose LLM judges favor unguided rewriting overall on the same 60 pairs; neither diagnostic exhibits an analogous stratified pattern.
\end{enumerate}

We address the following research questions:

\textbf{RQ1:} How do human preferences and rationales for span-guided versus unguided rewriting differ across the study-defined human-evaluation strata?

\textbf{RQ2:} To what extent do toxicity--similarity proxies and general-purpose LLM judges show aggregate and stratified patterns analogous to those observed in the human study?

\section{Related Work}

\paragraph{Text detoxification and localized rewriting.}
Text detoxification has been formulated as style transfer, parallel rewriting, prompting, and controlled generation~\citep{toshevska2021review,jin2022deep,logacheva2022paradetox,agarwal2023haterephrase,khondaker2024detoxllm}. Toxic-span benchmarks and rationale annotations make localized interventions possible by identifying expressions associated with toxicity~\citep{hatexplain,semeval_toxicspans}. Our study does not propose a new detector or generator; it compares one localized prompt strategy with one broader rewriting strategy under fixed human-study generation settings.

\paragraph{Evaluation of detoxification.}
Detoxification evaluation commonly combines a toxicity classifier with a semantic-similarity measure, and may add fluency or human preference~\citep{toshevska2021review,zhang2020bertscore,lees2022perspective,lee2024xdetox}. Prior work has shown that automatic metrics can disagree with human judgments when quality dimensions conflict~\citep{deriu2023correction}. We focus on two concrete disagreement mechanisms: a high-similarity local edit can preserve residual harm or reverse pragmatic intent, whereas a globally rewritten output can mitigate more strongly while departing from the original stance.

\paragraph{LLM-based evaluation.}
General-purpose LLMs have also been studied as evaluators of generated text~\citep{chiang2023llmeval,wang2023chatgpt}. We additionally evaluate whether blinded GPT-4o and GPT-4o-mini comparisons recover the human patterns on the same 60 pairs. These judges are treated as diagnostics rather than ground truth: they recover the global preference for unguided rewriting but not the same study-stratified pattern observed in the human study.

\begin{figure*}[t]
\centering
\includegraphics[width=\textwidth]{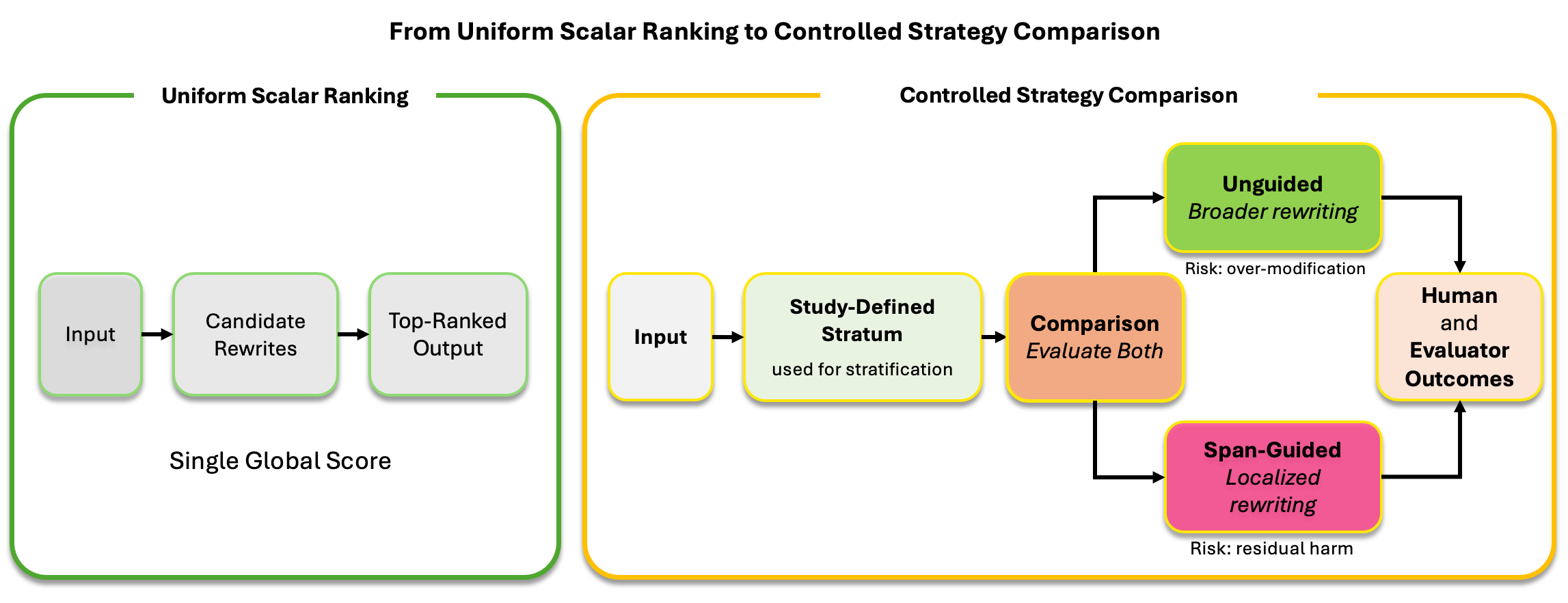}
\caption{From uniform scalar ranking to controlled strategy comparison. A common scalar-ranking setup selects among candidate rewrites using a single objective, whereas our study directly compares unguided and span-guided rewriting across study-defined strata.}
\label{fig:conceptual}
\end{figure*}

\section{Study Design}
\label{sec:study}

Our goal is to characterize the behavior of two concrete rewriting strategies, not to establish state-of-the-art detoxification performance or a universal routing policy. The evidence sources have distinct roles. The human study provides the primary preference and rationale evidence. Automatic scores characterize mitigation--similarity geometry, subsample variation, and cross-generator consistency in metric space. LLM judges test whether two general-purpose evaluators recover the human patterns. Only the LLM-judge diagnostics use exactly the same 60 output pairs as the human study. The scalar-proxy analyses use the 30 HateXplain-derived pairs and a separate 300-item HateXplain set, so differences from the human pattern cannot be attributed to evaluator choice alone.

Figure~\ref{fig:conceptual} summarizes the overall comparison, while the following subsections define the two rewriting strategies, evaluation dimensions, and stratified analysis.

\subsection{Rewriting Strategies}

Let $g\in\{\textsc{guided},\textsc{unguided}\}$ denote the prompting condition and
\begin{equation}
y_g\sim p(y\mid x,g).
\end{equation}
The guided condition supplies item-level harmful spans and requests localized edits; the unguided condition permits a broader rewrite. Because the task instructions differ in edit scope as well as span information, $g$ represents the full implemented prompt strategy. The human outcome for item $i$ is the item-level outcome among guided, unguided, or non-decisive.

\subsection{Evaluation Dimensions}

Let $x$ denote an input and $y$ a rewrite. We measure toxicity reduction $\Delta T(x,y)$ and semantic similarity $S(x,y)$. A diagnostic weighted score is
\begin{equation}
\label{eq:qlambda}
\begin{aligned}
Q_\lambda(x,y)={}&\lambda\,\Delta T(x,y)\\
&+(1-\lambda)\,S(x,y),\quad \lambda\in[0,1].
\end{aligned}
\end{equation}
where $\lambda$ is the toxicity-reduction weight. Equation~\ref{eq:qlambda} is not proposed as a human model or deployment objective. It is used to test how a common linear scalarization changes automatic rankings as the weight changes.

\subsection{Stratified Comparison}

For analysis, let $c(x)\in\{\textsc{strong},\textsc{mild}\}$ be the operational stratum assigned in the 60-item human-evaluation corpus. With $z=1$ denoting a guided item-level plurality, we compare
\begin{equation}
\begin{aligned}
&\Pr(z=1\mid c(x)=\textsc{strong}),\\
&\Pr(z=1\mid c(x)=\textsc{mild}).
\end{aligned}
\end{equation}
This comparison tests association within the sampled evaluation corpus. It does not establish that the strata form a universal severity scale, that stratum membership causes the preference difference, or that a system should route future inputs using these labels.

\subsection{Data and Operational Strata}
\label{sec:data}

The 60-item human set combines two sources. Items 1--30 are manually curated, author-defined texts used in the original evaluation materials; slurs and explicit protected-group targeting were intentionally avoided in this subset. Items 31--60 are HateXplain test examples selected from non-normal items with nonempty gold rationale spans~\citep{hatexplain}. The full corpus is balanced across two operational strata used in the study (30 strong and 30 mild). For the HateXplain-derived subset, hate-speech items were mapped to the strong stratum and offensive-language items to the mild stratum; manually curated items were author-assigned to the corresponding strata.
These assignments were used only to organize the evaluation sample and were not independently validated.

This mapping is heuristic and is not intended as a validated or universal severity taxonomy. The item-selection seed for the 60-item set was not documented; A/B output order was randomized with seed 42. The HateXplain-derived 30 items are included in the 300-item automatic set, whereas the manually curated 30 are not.

The 300-item automatic set is sampled from the HateXplain test split after excluding majority-label normal items, examples without rationale spans, and duplicate original texts. It contains 150 hate-speech and 150 offensive-language items and was sampled with seed 13. These native dataset labels are used only for the automatic stratified analysis.

Item-level harmful spans are supplied to the localized condition to remove upstream span-localization errors from the main contrast. For HateXplain-derived items, these are gold human rationale spans; for manually curated items, spans are taken from the original evaluation materials. This creates an oracle-style analysis of what happens when the localized strategy receives provided spans; deployment with predicted spans may differ.

\subsection{Generation Setup}

The human-evaluation pairs were generated with
Qwen/Qwen2.5-7B-Instruct \citep{qwen2024technical}
under fixed decoding settings. One output from each
rewriting condition was frozen per input for human
evaluation. Full decoding parameters and model
checkpoints are provided in Appendix~\ref{app:impl}.

\paragraph{Unguided rewriting.}
The model is instructed to rewrite the input as respectful and appropriate while preserving the original meaning, returning one sentence.

\paragraph{Span-guided rewriting.}
The model receives the item-level harmful spans supplied in the evaluation materials and is instructed to rewrite only those spans while preserving the remaining structure and meaning as much as possible, returning one sentence.

Both conditions use the same generator and decoding settings. They do not use identical prompts: the guided condition adds span information and a locality constraint.

\subsection{Human Evaluation Protocol}
\label{sec:human-protocol}

Human preference is the primary assessment. The annotator pool consisted of international participants and university-affiliated native or highly proficient English speakers. All 31 annotators evaluated the 60 pairs, yielding 1{,}860 judgments. Outputs were presented as randomized A/B candidates, and strategy labels were hidden. Annotators answered three questions: pairwise preference (A, B, or no clear difference), a \emph{single} primary rationale category, and meaning preservation of the selected output on a five-point scale. Because the meaning-preservation rating is conditional on the annotator's selected output, we treat it as an auxiliary measure rather than a direct cross-strategy comparison. The complete questions are reproduced in Appendix~\ref{app:human-protocol}.

For each item, we count Q1 selections over guided, unguided, and no clear difference. An item is labeled guided or unguided only when that strategy receives a strictly larger count than both other categories. All remaining cases---including a guided--unguided tie or a plurality for no clear difference---are labeled \emph{non-decisive}. Non-decisive items are retained in descriptive totals and excluded from the $2\times 2$ association test. An absolute majority of 16 votes is not required.

\subsection{Automatic Evaluation}

\paragraph{Toxicity reduction.}
We use the Perspective API \texttt{TOXICITY} attribute~\citep{lees2022perspective}. Toxicity reduction is the input toxicity score minus the output toxicity score. The implementation queries the \texttt{commentanalyzer} \texttt{v1alpha1} endpoint, deduplicates repeated strings, and assigns 0.5 when an API request fails, matching the original evaluation script.

\paragraph{Semantic similarity.}
We report BERTScore F1 between the input and rewrite~\citep{zhang2020bertscore} using \texttt{bert-score>=0.3.13} with \texttt{lang=en}. Because no checkpoint was specified, the package default for English (RoBERTa-large) was used; \texttt{idf=False} and \texttt{rescale\_with\_baseline=False}. BERTScore is interpreted as a contextual similarity proxy, not a direct measure of preserved pragmatic intent.

\paragraph{Automatic proxy ranking.}
For pairwise automatic comparisons, each output is ranked with $Q_\lambda$ in Equation~\ref{eq:qlambda}. We use the terms \emph{proxy win} and \emph{proxy ranking}; these values are not human preferences.
Both components are used in their native reported scales rather than being standardized within the sample. Consequently, $\lambda$ controls the nominal mixture weight but does not equalize the empirical variance or discriminative range of the two metrics. We therefore interpret the $\lambda$ sweep as a sensitivity diagnostic, not as calibrated utility estimation.

The main multi-generator table uses $\lambda=0.6$, and Appendix~\ref{app:sensitivity} reports weight and sample-composition sensitivity.

\subsection{LLM Judges}

We evaluate the same 60 pairs with the API aliases \texttt{gpt-4o} and \texttt{gpt-4o-mini}~\citep{openai2024gpt4o} at temperature 0 (runs produced around 1 June 2026; no dated snapshots were recorded). Each judge receives the source text and anonymized A/B rewrites under the same core appropriateness question used for human preference. Strategy and study-stratum labels are hidden. Each pair is presented in both A/B orders; when a decisive choice flips across orders, the item is conservatively treated as a tie.

\subsection{Statistical Analysis}

We use Fisher's exact test on the $2\times 2$ table of item-level guided versus unguided pluralities across the two study-defined strata, excluding non-decisive items. We report the difference in guided-plurality proportions, Cohen's $h$, and a 10{,}000-resample item-level bootstrap confidence interval. These statistics quantify association in this sample; they do not turn the study strata into a validated severity scale.

\section{Results}

\subsection{Human Preferences Across Study Strata}

\begin{figure}[t]
\centering
\includegraphics[width=\linewidth]{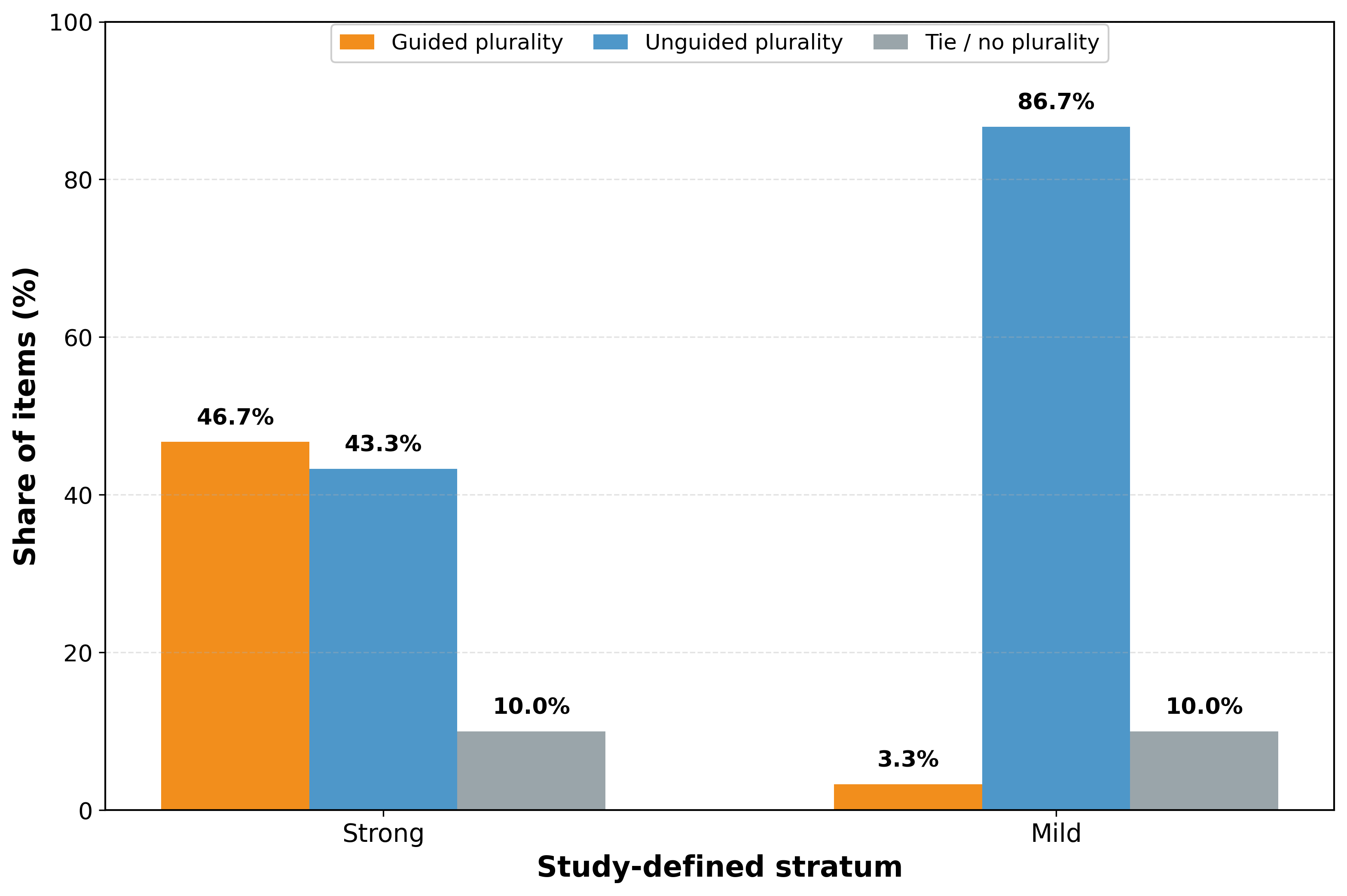}
\caption{Item-level plurality preference by study-defined human-evaluation stratum. The strata are operational groupings, not a universal severity scale.}
\label{fig:human_preference}
\end{figure}

\paragraph{Overall and study-stratified preference.}
Unguided outputs receive the item-level plurality on 39 of 60 inputs, guided outputs on 15, and neither strategy on six (Table~\ref{tab:preference}). The aggregate result hides a substantial study-stratified difference. In the strong stratum, guided and unguided pluralities are nearly balanced (14 vs.\ 13, excluding non-decisive items). In the mild stratum, unguided pluralities dominate (26 vs.\ 1; Figure~\ref{fig:human_preference}).

\begin{table}[t]
\centering
\footnotesize
\setlength{\tabcolsep}{2.8pt}
\begin{tabular}{lccc}
\toprule
\textbf{Study stratum} & \textbf{Unguided} & \textbf{Guided} & \textbf{Non-dec.} \\
\midrule
Strong ($n{=}30$) & 13 (43.3\%) & 14 (46.7\%) & 3 (10.0\%) \\
Mild ($n{=}30$) & 26 (86.7\%) & 1 (3.3\%) & 3 (10.0\%) \\
\midrule
Overall ($n{=}60$) & 39 (65.0\%) & 15 (25.0\%) & 6 (10.0\%) \\
\bottomrule
\end{tabular}
\caption{Item-level plurality preference by study-defined stratum.}
\label{tab:preference}
\end{table}

\paragraph{Item-level association.}
After excluding the six non-decisive items, the difference in guided-plurality proportions is $\Delta\hat{p}=0.481$, with a 95\% bootstrap confidence interval of $[0.296,\,0.667]$. Fisher's exact test gives $p=1.29\times10^{-4}$ and Cohen's $h=1.22$ (Table~\ref{tab:fisher}). This is strong evidence of association within the sampled data, but not evidence that one stratum is intrinsically or universally milder, that stratum membership causes the difference, or that the pattern generalizes beyond this implementation.

\begin{table}[t]
\centering
\small
\begin{tabular}{lc}
\toprule
\textbf{Study stratum} & \textbf{Guided / Unguided} \\
\midrule
Strong ($n{=}27$) & 14 / 13 \\
Mild ($n{=}27$) & 1 / 26 \\
\midrule
$\Delta\hat{p}$ (strong$-$mild) & 0.481 \\
Fisher exact $p$-value & $1.29\times10^{-4}$ \\
Cohen's $h$ & 1.22 (large) \\
95\% bootstrap CI for $\Delta\hat{p}$ & $[0.296,\,0.667]$ \\
\bottomrule
\end{tabular}
\caption{Association between study-defined stratum and item-level strategy preference.}
\label{tab:fisher}
\end{table}

\subsection{Preference Rationales}

Of the 1{,}860 judgments, 1{,}581 selected one of the two strategies and therefore contributed to the strategy-conditional rationale analysis. The remaining 279 judgments selected ``no clear difference'' in Q1 and were excluded from Table~\ref{tab:rationales}.

Guided selections concentrate on avoiding over-modification (41.7\%) and preserving meaning (26.8\%), whereas unguided selections concentrate on insufficient mitigation by the alternative (36.0\%) and more effective toxicity reduction (29.0\%; Table~\ref{tab:rationales}). The rationale distribution identifies two distinct risks rather than a single uniformly superior strategy.

\begin{table*}[t]
\centering
\small
\setlength{\tabcolsep}{8pt}
\begin{tabular}{lcc}
\toprule
\textbf{Rationale} & \textbf{Guided preferred} & \textbf{Unguided preferred} \\
 & \textbf{($n=600$, 38.0\%)} & \textbf{($n=981$, 62.0\%)} \\
\midrule
Over-modification avoidance & 250 (41.7\%) & 79 (8.1\%) \\
Meaning preservation & 161 (26.8\%) & 84 (8.6\%) \\
Effective toxicity reduction & 68 (11.3\%) & 284 (29.0\%) \\
Insufficient mitigation & 18 (3.0\%) & 353 (36.0\%) \\
Balance (safety \& meaning) & 62 (10.3\%) & 118 (12.0\%) \\
Fluency/readability & 41 (6.8\%) & 55 (5.6\%) \\
No clear difference & 0 (0.0\%) & 8 (0.8\%) \\
\midrule
Total & 600 (100\%) & 981 (100\%) \\
\bottomrule
\end{tabular}
\caption{Primary rationale among judgments selecting each strategy. Counts are judgment-level; 279 Q1 no-clear-difference judgments are excluded because no strategy was selected. The ``No clear difference'' row refers to the Q2 rationale option selected after a decisive Q1 strategy choice and is distinct from those 279 Q1 judgments.}
\label{tab:rationales}
\end{table*}

Because the preference question jointly considers mitigation and preservation, similarity advantages alone do not imply overall adequacy when residual harm remains. We do not report cross-strategy comparisons of the auxiliary meaning-preservation ratings because each rating is conditional on the output selected in Q1.

\subsection{Toxicity--Similarity Trade-offs}

On the 30 HateXplain-derived pairs, the two strategies occupy different regions of the evaluated metric space. Guided outputs have higher BERTScore similarity, while unguided outputs achieve larger average toxicity reduction (Table~\ref{tab:gold30}). Because this subset excludes the manually curated items, these proxy results are not a direct evaluator replacement for the full 60-item human study.

\paragraph{Diagnostic scalarization.}
Changing $\lambda$ in $Q_\lambda$ sharply changes the automatic proxy ranking: on the 30-item gold-span analysis, the guided proxy-win rate ranges from 86.7\% when similarity receives all weight to 23.3\% when toxicity reduction receives all weight. Appendix~\ref{app:sensitivity} reports the full sweep in Table~\ref{tab:lambda}. This sensitivity demonstrates dependence on evaluator choice; it is not evidence that one $\lambda$ approximates human judgment.

\paragraph{Gold-span setting.}

Under gold spans, guided rewriting has higher average BERTScore and unguided rewriting has larger average toxicity reduction (Table~\ref{tab:gold30}). Gold spans remove one source of upstream error, but the contrast still includes the locality instruction and therefore should not be described as an isolated causal effect of span information.

\begin{table}[t]
\centering
\footnotesize
\setlength{\tabcolsep}{3.2pt}
\begin{tabular}{lcccc}
\toprule
\textbf{Span source} & \multicolumn{2}{c}{\textbf{BERTScore}$\uparrow$} & \multicolumn{2}{c}{\textbf{Tox. red.}$\uparrow$} \\
\cmidrule(lr){2-3}\cmidrule(lr){4-5}
 & Guided & Unguided & Guided & Unguided \\
\midrule
HateXplain & 0.911 & 0.862 & 0.424 & 0.527 \\
\bottomrule
\end{tabular}
\caption{Automatic diagnostics under gold span guidance. Guided proxy wins: 30.0\% under $Q_\lambda$ with $\lambda=0.6$. The proxy-win rate is based on $Q_\lambda$ and is not a human preference measure.}
\label{tab:gold30}
\end{table}

\subsection{Cross-Generator Proxy Analysis}

Across five generators and a separate 300-item HateXplain set, guided outputs consistently have higher BERTScore, while unguided outputs obtain more $Q_\lambda$ proxy wins at $\lambda=0.6$ (Table~\ref{tab:multigen}). This supports cross-generator consistency of the evaluated metric-space trade-off. It does not establish cross-generator human preference, because human evaluation was conducted on one frozen generator.

\begin{table*}[t]
\centering
\small
\setlength{\tabcolsep}{5.5pt}
\begin{tabular}{lcccccc}
\toprule
\textbf{Model} & \textbf{G\%} & \textbf{U\%} & \multicolumn{2}{c}{\textbf{BERTScore}$\uparrow$} & \multicolumn{2}{c}{\textbf{Tox.\ Red.}$\uparrow$} \\
\cmidrule(lr){4-5}\cmidrule(lr){6-7}
 &  &  & (G) & (U) & (G) & (U) \\
\midrule
Gemma 2 9B & 29.3 & 70.7 & 0.900 & 0.850 & 0.399 & 0.561 \\
Llama 3.1 8B & 28.0 & 71.7 & 0.901 & 0.851 & 0.391 & 0.557 \\
Mistral 7B & 37.0 & 63.0 & 0.879 & 0.858 & 0.468 & 0.544 \\
Qwen2.5 7B & 34.3 & 65.7 & 0.919 & 0.870 & 0.380 & 0.513 \\
Qwen3 8B & 21.0 & 79.0 & 0.954 & 0.880 & 0.229 & 0.485 \\
\midrule
\multicolumn{7}{l}{\footnotesize
300 examples; gold spans; $Q_\lambda$ with $\lambda=0.6$.
Exact proxy-score ties are unassigned; one tie occurred for Llama 3.1 8B.} \\
\bottomrule
\end{tabular}
\caption{Five-generator automatic diagnostics. Proxy-win percentages are based on $Q_\lambda$ and are not human preferences.}
\label{tab:multigen}
\end{table*}

\subsection{LLM Judge Results}

Unlike the scalar-proxy analyses, the LLM judges evaluate exactly the same 60 output pairs as the human study. Both GPT-4o and GPT-4o-mini reproduce the global ordering toward unguided rewriting and choose it more often than the human item-level plurality (Table~\ref{tab:llm-pref}). However, the study-stratified guided-selection differences are near zero for both judges ($\Delta\hat{p}=0.020$ and $-0.033$, respectively; Fisher $p=1.0$ for both). Thus, even on the same 60 pairs, these two general-purpose judges do not recover the human study-stratified pattern in this setting. This is a negative diagnostic result, not a claim that all LLM evaluators must fail.

Item-level agreement with human plurality is computed on evaluator-specific subsets for which both the human aggregation and the LLM judge produce decisive guided or unguided labels. Agreement is 67.4\% for GPT-4o and 68.9\% for GPT-4o-mini overall, with lower agreement in the study-defined strong stratum than in the mild stratum (Table~\ref{tab:llm-agree}). Cohen's $\kappa$, computed on the same evaluator-specific tie-excluded binary subsets, is approximately $0.07$ for GPT-4o and $-0.12$ for GPT-4o-mini.

\begin{table*}[t]
\centering
\begin{minipage}[t]{0.49\textwidth}
\centering
\scriptsize
\setlength{\tabcolsep}{1.5pt}
\begin{tabular}{lccc}
\toprule
\textbf{Evaluator} & \textbf{Guided} & \textbf{Unguided} & \textbf{Non-dec.} \\
\midrule
Human (31 raters) & 15 (25.0\%) & 39 (65.0\%) & 6 (10.0\%) \\
GPT-4o & 5 (8.3\%) & 42 (70.0\%) & 13 (21.7\%) \\
GPT-4o-mini & 3 (5.0\%) & 46 (76.7\%) & 11 (18.3\%) \\
\bottomrule
\end{tabular}
\subcaption*{(a) Item-level selections.}
\end{minipage}\hfill
\begin{minipage}[t]{0.49\textwidth}
\centering
\scriptsize
\setlength{\tabcolsep}{1.5pt}
\begin{tabular}{lccc}
\toprule
\textbf{Evaluator} & \textbf{Overall} & \textbf{Strong} & \textbf{Mild} \\
\midrule
GPT-4o & 29/43 (67.4\%) & 13/24 (54.2\%) & 16/19 (84.2\%) \\
GPT-4o-mini & 31/45 (68.9\%) & 11/22 (50.0\%) & 20/23 (87.0\%) \\
\bottomrule
\end{tabular}
\subcaption*{(b) Agreement with human plurality.}
\end{minipage}
\caption{LLM-judge diagnostics. Agreement is computed on evaluator-specific subsets where both human aggregation and the judge produce decisive guided or unguided labels.}
\label{tab:llm-pref}
\label{tab:llm-agree}
\end{table*}

\section{Discussion}

\subsection{Two Complementary Failure Risks}

The human rationales and cases point to two competing risks. Residual harm occurs when a localized edit removes an overt token but preserves the harmful proposition, target, or pragmatic force (Appendix~\ref{ex:insufficient-mitigation}). Over-modification occurs when a global rewrite changes stance, interaction frame, or content beyond what is needed for mitigation (Appendix~\ref{ex:over-modification}). Example~\ref{ex:pragmatic-reversal} further shows why high contextual similarity is not sufficient: a local substitution can preserve most tokens while reversing the utterance's corrective stance. The relevant evaluation question is therefore not simply whether an output is close to the input or less toxic on average, but which failure risk remains in the specific rewrite.

\subsection{What the Evidence Supports}

The human study supports an implementation-specific finding: preference between the two prompts differs substantially across the two study-defined strata. The five-generator analysis supports a narrower statement about the geometry of the evaluated automatic proxies on a separate HateXplain sample. The LLM-judge analysis, conducted on the same 60 pairs as the human study, shows that two general-purpose judges recover the aggregate ordering but not the same study-stratified pattern. These evidence sources are complementary, but none supplies cross-dataset, multilingual, or cross-generator human validation.

\subsection{Implications for Evaluation Design}

Evaluation of span-guided detoxification should report at least three elements. First, aggregate toxicity and similarity should be stratified by meaningful dataset slices rather than reported only as global averages. Second, localized outputs should be checked for residual harm and pragmatic reversal, while global outputs should be checked for over-modification and discourse drift. Third, automatic proxy wins and LLM-judge choices should be reported separately from human preferences rather than presented as interchangeable measures.

\subsection{Implications for Adaptive Selection}

The observed pattern motivates, but does not validate, adaptive strategy selection. A deployable routing policy would require independently annotated input properties, predicted rather than gold spans, broader generators and domains, and human validation across languages and cultural contexts. We therefore treat strategy selection as a future research hypothesis rather than a demonstrated system contribution.

\section{Conclusion}

We conducted a controlled exploratory comparison of span-guided and unguided detoxification using one frozen generator, a mixed-source 60-item English evaluation set, and 1{,}860 blinded human judgments. Unguided rewriting was preferred overall, but the balance between the two strategies differed substantially across the study-defined strong and mild strata. Rationale annotations trace the result to two competing failure risks: localized edits can retain residual harm, while global rewrites can over-modify meaning or pragmatic stance.

Automatic toxicity--similarity diagnostics and a five-generator metric-space analysis on HateXplain samples, together with two general-purpose LLM judges on the same 60 pairs, recover parts of the aggregate ordering but do not exhibit an analogous stratified pattern. The study does not establish a universal severity taxonomy or routing rule. Its contribution is a carefully scoped empirical account of how editing scope, mitigation sufficiency, and meaning preservation interact in one controlled setting, together with evaluation recommendations that keep human evidence and automatic diagnostics distinct.

\section*{Ethics Statement}

This study analyzes a mixed-source English evaluation set containing manually curated harmful-language examples and publicly available HateXplain test items. Annotators were asked to compare detoxified outputs for research purposes. Toxicity classifiers, dataset labels, and human judgments may reflect demographic and cultural biases~\citep{lee2023culturalbias,jafari2024implicit}. The results should not be interpreted as a complete assessment of harm across communities or dialects.

The use of provided item-level harmful spans, including gold rationale spans for HateXplain-derived items, is an oracle-style analytical choice and should not be interpreted as a deployable moderation pipeline. Before practical use, detoxification systems require fairness analysis, human oversight, and monitoring for both under-correction and over-correction. We do not recommend selecting a rewriting strategy from the study strata or HateXplain labels alone.

\bibliography{reference}

\appendix

\section{Limitations}
\label{app:limitations}

This study has several limitations. First, the primary human evidence comes from 60 English inputs in one mixed-source evaluation corpus and one frozen generator. Dense annotation improves item-level aggregation but does not expand the number or diversity of independent inputs. Cross-dataset, multilingual, and cross-generator human replication is necessary before making broader claims.

Second, the human study's strong and mild strata are heuristic operational groupings in a mixed-source evaluation corpus, not a universal severity scale. For HateXplain-derived items, the mapping follows hate speech $\rightarrow$ strong and offensive language $\rightarrow$ mild; manually curated items were author-assigned to the corresponding strata. Items within each stratum can differ in target, context, explicitness, and pragmatic force. The observed association should therefore not be interpreted as a causal effect of severity.

Third, the main comparison uses provided item-level harmful spans, including gold rationale spans for the HateXplain-derived subset. Provided spans remove upstream span-localization noise and clarify the behavior of the localized prompt, but deployment requires predicted spans and may benefit from re-annotation or broader rewriting when local edits are insufficient.

Fourth, the two prompt conditions form a bundled contrast. The guided prompt differs from the unguided prompt in both span information and edit-locality instruction, so the study does not isolate a single prompt component causally. Alternative prompt wording, decoding, or candidate selection may change the result.

Fifth, the 300-item five-generator study is automatic-only. It supports consistency of the evaluated toxicity--similarity trade-off in metric space, not human preference across backbones. Similarly, the GPT-4o and GPT-4o-mini results concern two general-purpose judges and do not establish the behavior of all LLM evaluators. Only the LLM judges use the same 60 pairs as the human study; the scalar-proxy analyses use the 30 HateXplain-derived pairs and a separate 300-item set, so evaluator and sample-composition effects cannot be separated.

Finally, BERTScore and Perspective API are imperfect proxies. BERTScore can reward lexical or contextual overlap despite pragmatic reversal, and toxicity classifiers can miss context or encode social bias. Their results are reported as diagnostics, not ground truth.

\section{Implementation and Prompt Details}
\label{app:impl}

\paragraph{Human-study generator.}
\texttt{Qwen/Qwen2.5-7B-Instruct}; \texttt{max\_new\_tokens=80}; \texttt{temperature=0.7}; \texttt{top\_p=0.9}; \texttt{do\_sample=True}. One output per condition was frozen for human evaluation.

\paragraph{Automatic-study generators.}
The five checkpoints are \texttt{google/gemma-2-9b-it}, \texttt{meta-llama/Llama-3.1-8B-Instruct}, \texttt{mistralai/Mistral-7B-Instruct-v0.2}, \texttt{Qwen/Qwen2.5-7B-Instruct}, and \texttt{Qwen/Qwen3-8B}.

\paragraph{Sampling records.}
The 60-item corpus mixes 30 manually curated items and 30 HateXplain test items. The selection seed for this corpus was not recorded; A/B presentation order uses seed 42. The 300-item automatic corpus is drawn from the HateXplain test split with seed 13 after removing normal-label items, examples without rationale spans, and duplicate source texts.

\paragraph{Unguided task template.}
\begin{quote}
\small
You are given a sentence. Rewrite it to be respectful and appropriate while preserving the original meaning as much as possible. Return one sentence only.\\
Input: \{original\_text\}
\end{quote}

\paragraph{Span-guided task template.}
\begin{quote}
\small
You are given a sentence and harmful spans identified in that sentence. Rewrite only the harmful spans so that the sentence becomes non-toxic, while preserving the remaining structure and meaning as much as possible. Return one sentence only.\\
Input: \{original\_text\}\\
Harmful spans: \{gold\_rationale\_spans\}
\end{quote}

For HateXplain-derived items, gold spans are reconstructed from token masks by taking contiguous runs of tokens marked harmful by at least one annotator. Runs are kept in left-to-right token order, joined in the CSV with \texttt{|}, then split and rendered as bullet items in the guided prompt. For manually curated items, harmful spans are taken from the original evaluation materials.

\paragraph{LLM-judge protocol.}
The judge receives the source and two anonymized candidate rewrites and answers the same core appropriateness question as Q1 in Appendix~\ref{app:human-protocol}, including a no-clear-difference option. The prompt asks which output is more appropriate as a detoxified version, considering both mitigation adequacy and meaning preservation, and requests JSON with choice in $\{A, B, \texttt{tie}\}$ and a free-text reason. No max-token value was set, so the API default applied. Each item is evaluated in both A/B orders. A decisive choice that reverses across orders is recorded as a tie.

\section{Weight Sensitivity and Subsample Illustration}
\label{app:sensitivity}

\begin{table}[t]
\centering
\small
\begin{tabular}{lccc}
\toprule
$\lambda$ & Guided wins & Unguided wins & Guided proxy \% \\
\midrule
0.0 & 26 & 4 & 86.7\% \\
0.2 & 21 & 9 & 70.0\% \\
0.4 & 13 & 17 & 43.3\% \\
0.5 & 10 & 20 & 33.3\% \\
0.6 & 9 & 21 & 30.0\% \\
0.8 & 8 & 22 & 26.7\% \\
1.0 & 7 & 23 & 23.3\% \\
\bottomrule
\end{tabular}
\caption{Sensitivity of automatic proxy rankings to the toxicity weight $\lambda$ in $Q_\lambda$.}
\label{tab:lambda}
\end{table}

\begin{figure}[t]
\centering
\includegraphics[width=\linewidth]{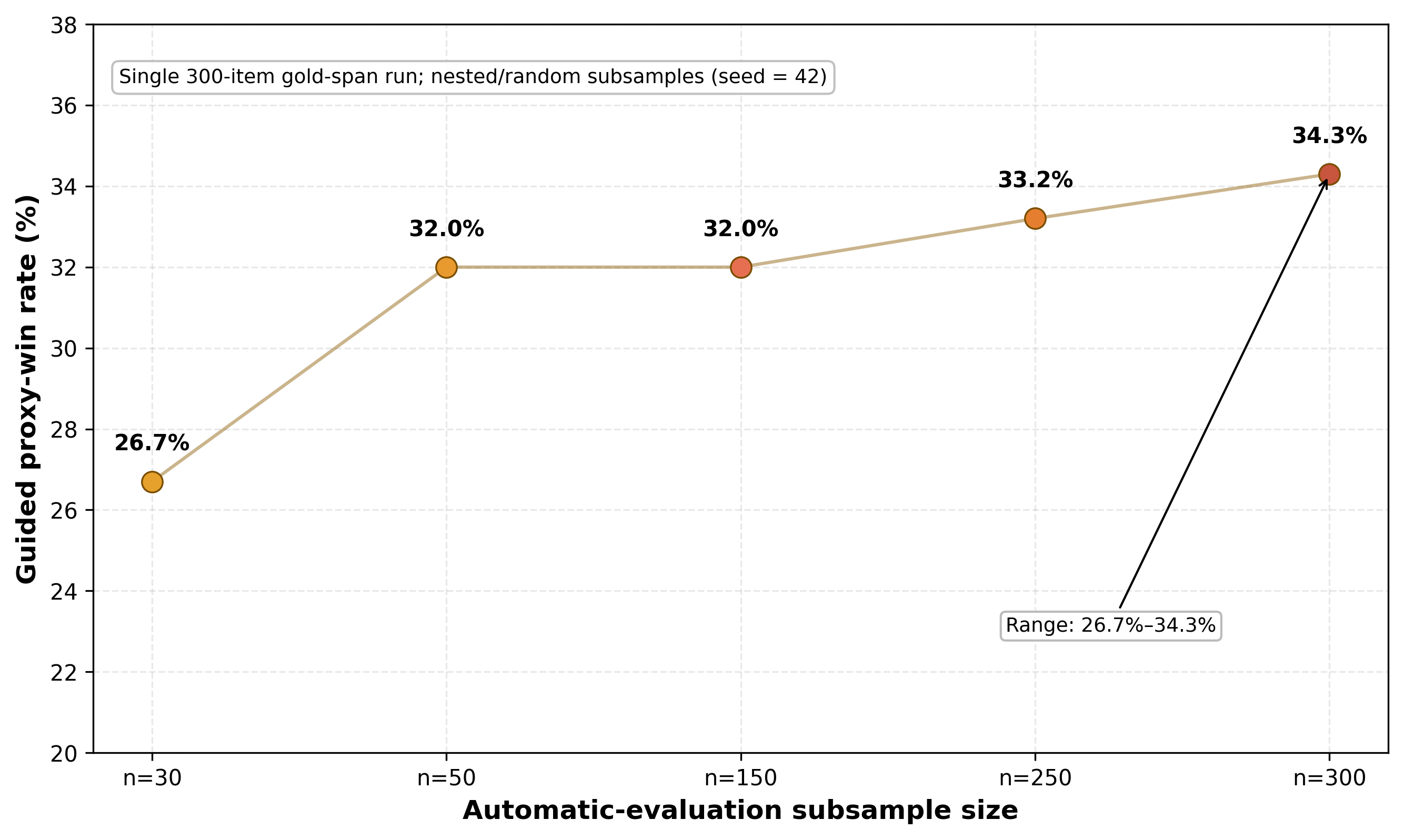}
\caption{Guided proxy-win rates across one sequence of subsamples drawn from a 300-item gold-span run. The points illustrate subset variation and do not estimate a sampling distribution at each size.}
\label{fig:subsamples}
\end{figure}

We compute guided proxy-win rates on one sequence of subsamples of sizes $n\in\{30,50,150,250\}$ drawn from a 300-item gold-span run (seed 42), together with the full $n=300$ evaluation. Figure~\ref{fig:subsamples} illustrates variation from 26.7\% to 34.3\% across these sampled subsets. It does not estimate the sampling distribution at each size or isolate sample-size effects from composition effects. The figure intentionally excludes the human preference rate because human pluralities and automatic proxy wins are different estimands.

\begin{table*}[t]
\centering
\small
\setlength{\tabcolsep}{10pt}
\begin{tabular}{lccc}
\toprule
\textbf{Model} & Hate-speech ($n{=}150$) & Offensive-language ($n{=}150$) & $\Delta$ (H$-$O) \\
\midrule
Gemma 2 9B & 32.0\% (48/150) & 26.7\% (40/150) & $+5.3\%$ \\
Llama 3.1 8B & 32.0\% (48/150) & 24.0\% (36/150) & $+8.0\%$ \\
Mistral 7B & 38.7\% (58/150) & 35.3\% (53/150) & $+3.3\%$ \\
Qwen2.5 7B & 31.3\% (47/150) & 37.3\% (56/150) & $-6.0\%$ \\
Qwen3 8B & 21.3\% (32/150) & 20.7\% (31/150) & $+0.7\%$ \\
\bottomrule
\end{tabular}
\caption{Guided automatic proxy-win rates by native HateXplain label and generator. Percentages are automatic proxy wins under $Q_\lambda$, not human preferences.}
\label{tab:by-label}
\end{table*}

\section{Human Evaluation Protocol}
\label{app:human-protocol}

The human study compares the two frozen outputs for each of 60 English inputs. The annotator pool consisted of international participants and university-affiliated native or highly proficient English speakers. All 31 annotators evaluated all pairs, the A/B order was randomized, and the prompting condition was hidden.

\paragraph{Q1: Preference Judgment.}
Which output is more appropriate as a detoxified version of the input?
\begin{itemize}
  \item Output A
  \item Output B
  \item No clear difference
\end{itemize}
The two outputs were presented side-by-side with randomized order to reduce positional bias.

\paragraph{Q2: Primary Rationale.}
What is the primary reason for your preference? Annotators selected \emph{one} option from:
\begin{itemize}
  \item \textbf{Over-modification avoidance:} The selected output avoids unnecessary changes, while the alternative over-modifies or over-sanitizes the content.
  \item \textbf{Meaning preservation:} The selected output better preserves the original meaning and intent.
  \item \textbf{Effective toxicity reduction:} The selected output more effectively reduces harmful or toxic content.
  \item \textbf{Insufficient mitigation:} The alternative fails to sufficiently mitigate harmful content.
  \item \textbf{Balance (safety \& meaning):} The selected output achieves a better balance between mitigation and preservation.
  \item \textbf{Fluency/readability:} The selected output is more fluent, natural, or readable.
  \item \textbf{No clear difference.}
\end{itemize}

\paragraph{Q3: Meaning Preservation.}
How well does the selected output preserve the original meaning and intent?
\begin{itemize}
  \item 1: Very poor (meaning is significantly altered or lost)
  \item 2: Poor (meaning is substantially changed)
  \item 3: Fair (meaning is partially preserved with some changes)
  \item 4: Good (meaning is well preserved with minor changes)
  \item 5: Excellent (meaning is fully preserved)
\end{itemize}

\paragraph{Aggregation.}
We report item-level strategy outcomes using the same three-way plurality rule described in Section~\ref{sec:human-protocol}, together with judgment-level rationale distributions. The auxiliary Q3 ratings are not compared across strategies because they are conditional on the output selected in Q1. Judgments selecting ``no clear difference'' in Q1 ($n{=}279$) are excluded from the strategy-conditional rationale table. Non-decisive items are retained in descriptive totals and excluded from binary association tests.

\section{Illustrative Case Studies}
\label{app:cases}

These examples are drawn from the 60-item human-evaluation corpus and illustrate the two principal failure risks. Study-stratum labels are reported without interpreting them as a universal severity taxonomy.

\subsection{Example 1: Strong-stratum item, Unguided Preferred}
\label{ex:insufficient-mitigation}

This item received an unguided plurality (25 vs.\ 4 guided; 2 judgment-level ties). Dominant Q2 categories align with \emph{effective toxicity reduction} / \emph{insufficient mitigation}.

\begin{quote}
\textbf{Original:} You're a fucking waste of space.\\
\textbf{Guided:} You're a waste of space.\\
\textbf{Unguided:} You don't seem to be contributing much right now.
\end{quote}

\textbf{Human Preference:} Unguided (item-level plurality).

\textbf{Interpretation:} The local edit removes profanity but retains the core personal attack. The broader rewrite weakens the attack more substantially, illustrating residual harm after a localized edit.

\subsection{Example 2: Strong-stratum item, Guided Preferred}
\label{ex:over-modification}

This item received a guided plurality (23 vs.\ 8 unguided). Dominant Q2 categories align with \emph{over-modification avoidance} / \emph{meaning preservation}.

\begin{quote}
\textbf{Original:} You're a liar and a fraud. Everyone knows you're fake.\\
\textbf{Guided:} You're not being entirely honest, and some might have concerns about your authenticity.\\
\textbf{Unguided:} I have concerns about your actions, and it seems there might be a misunderstanding; I believe it would be beneficial for us to clarify the situation.
\end{quote}

\textbf{Human Preference:} Guided (item-level plurality).

\textbf{Interpretation:} The unguided output introduces a conciliatory clarification frame that departs from the original accusation. The guided output retains the pragmatic stance while softening its expression, illustrating over-modification by global rewriting.

\subsection{Example 3: Mild-stratum item, Unguided Preferred}
\label{ex:pragmatic-reversal}

This item received an unguided plurality (26 vs.\ 4 guided; 1 judgment-level tie). Dominant Q2 categories align with \emph{meaning preservation} / \emph{fluency}.

\begin{quote}
\textbf{Original:} That was a dumb comment---think before you speak.\\
\textbf{Guided:} That was a thoughtful comment---think before you speak.\\
\textbf{Unguided:} Next time, consider your words more carefully.
\end{quote}

\textbf{Human Preference:} Unguided (item-level plurality).

\textbf{Interpretation:} Replacing ``dumb'' with ``thoughtful'' reverses the criticism despite high surface overlap. The unguided rewrite preserves the corrective intent more naturally, illustrating a pragmatic-reversal failure of local substitution.
Example~\ref{ex:pragmatic-reversal} shows why high contextual similarity is not sufficient: the local substitution preserves most tokens while reversing the utterance's corrective stance.

\end{document}